\documentclass[lettersize,journal]{IEEEtran}
\usepackage{amsmath,amsfonts}
\usepackage{algorithmic}
\usepackage{algorithm}
\usepackage{array}
\usepackage[caption=false,font=normalsize,labelfont=sf,textfont=sf]{subfig}
\usepackage{textcomp}
\usepackage{stfloats}
\usepackage{url}
\usepackage{verbatim}
\usepackage{graphicx}
\usepackage{cite}
\usepackage{tabularx}
\usepackage{booktabs}
\usepackage{color,colortbl}
\usepackage{hyperref}

\newcolumntype{L}[1]{>{\raggedright\arraybackslash}m{#1}}
\newcolumntype{C}[1]{>{\centering\arraybackslash}p{#1}}

\begin{document}

\title{DiffFashion: Reference-based Fashion Design with Structure-aware Transfer by Diffusion Models}

\author{Shidong Cao$^*$, Wenhao Chai$^*$, Shengyu Hao, Yanting Zhang, Hangyue Chen$^\dagger$, and Gaoang Wang$^\dagger$,~\IEEEmembership{Member,~IEEE}

\thanks{$^*$ Equal contribution.}
\thanks{$^\dagger$ Corresponding author: Hangyue Chen, and Gaoang Wang.}
\thanks{Shidong Cao, Wenhao Chai, and Shengyu Hao  are with the Zhejiang University-University of Illinois Urbana-Champaign Institute, Zhejiang University, China (e-mail:
22271126@zju.edu.cn,
wenhaochai.19@intl.zju.edu.cn,
shengyuhao@zju.edu.cn).}
\thanks{
Yanting Zhang is with the Donghua University, China (e-mail: ytzhang@dhu.edu.cn).}
\thanks{Hangyue Chen is with the Hangzhou Dianzi University, China (e-mail: chy@hdu.edu.cn).}
\thanks{Gaoang Wang is with the Zhejiang University-University of Illinois Urbana-Champaign Institute, and College of Computer Science and Technology, Zhejiang University, China (e-mail: gaoangwang@intl.zju.edu.cn).}
}



\maketitle

\begin{abstract}
Image-based fashion design with AI techniques has attracted increasing attention in recent years. 
We focus on a new fashion design task, where we aim to transfer a reference appearance image onto a clothing image while preserving the structure of the clothing image. It is a challenging task since there are no reference images available for the newly designed output fashion images. Although diffusion-based image translation or neural style transfer (NST) has enabled flexible style transfer, it is often difficult to maintain the original structure of the image realistically during the reverse diffusion, especially when the referenced appearance image greatly differs from the common clothing appearance.
To tackle this issue, we present a novel diffusion model-based unsupervised structure-aware transfer method to semantically generate new clothes from a given clothing image and a reference appearance image. 
In specific, we decouple the foreground clothing with automatically generated semantic masks by conditioned labels. And the mask is further used as guidance in the denoising process to preserve the structure information. Moreover, we use the pre-trained vision Transformer (ViT) for both appearance and structure guidance. Our experimental results show that the proposed method outperforms state-of-the-art baseline models, generating more realistic images in the fashion design task. Code and demo can be found at \href{https://github.com/Rem105-210/DiffFashion}{https://github.com/Rem105-210/DiffFashion}.
\end{abstract}

%"The utilization of AI techniques in image-based fashion design has been the subject of growing interest in recent years. Our focus is directed towards an innovative fashion design challenge where the objective is to transfer the aesthetic of a reference image onto a clothing image while safeguarding the latter's structural integrity. This poses a formidable task as there is a lack of reference images for the newly designed fashion outputs. Although current diffusion-based image translation or neural style transfer methods allow for adaptable style transfer, it remains a challenge to retain the original image structure in a convincing manner, particularly when the reference image differs significantly from the typical appearance of clothing. To surmount this issue, we introduce a novel unsupervised structure-aware transfer approach based on diffusion modeling, which semantically creates new garments from a given clothing image and reference appearance image. The foreground clothing is separated through the utilization of automatically generated semantic masks, conditioned by labels. These masks are then employed as a guide in the denoising process to preserve the structural information. Additionally, we employ a pre-trained vision Transformer for both appearance and structure guidance. Our experiments demonstrate that our proposed method outperforms existing state-of-the-art models, producing more lifelike images in the fashion design task."


\section{Introduction} %这intro的逻辑能先列出来吗
\begin{figure*}[!t]
\centering
\includegraphics[width=1.0\linewidth]{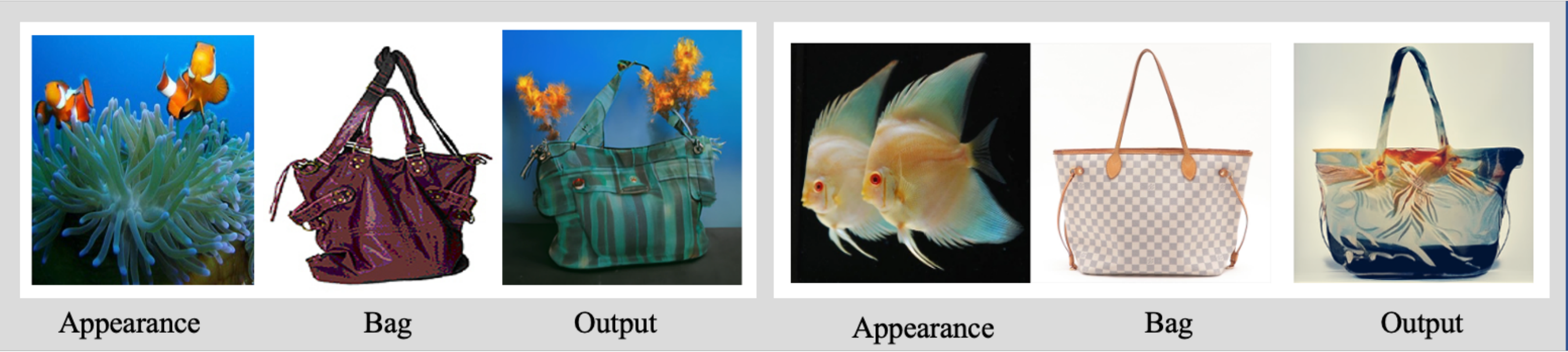}
\caption{Two examples of a reference-based fashion design task. For a given image pair, \textit{i.e.}, a bag and a referenced appearance image, our method can generate a new image with appearance similarity to the appearance image and structure similarity to the bag image.}
\label{example}
\end{figure*}

Image-based fashion design with artificial intelligence (AI) techniques \cite{ganesan2017fashioning,yan2022toward,sbai2018design,kim2019style,yan2022toward_tmm,zhou2022coutfitgan} has attracted increasing attention in recent years.
There is a growing expectation that AI can provide inspiration for human designers to create new fashion designs. One of the emerging tasks in fashion design is to add specific texture elements from non-fashion domain images into clothing images to create new fashions. For example, given a clothing image, a designer may want to generate a new clothes design with the appearance of another domain object as a reference, as shown in Fig.~\ref{example}.

Generative adversarial network (GAN)-based methods \cite{yan2022toward,yuan2020garment,cui2018fashiongan} can be adopted in the common fashion design tasks to generate new clothes. However, GAN-based methods can hardly have good control over the appearance and shape of clothes when transferring from non-fashion domain images.
Recently, diffusion models \cite{ho2020denoising,song2020denoising,song2020score} have been widely explored due to the realism and diversity of their results, and have been applied in various generative areas, such as text image generation \cite{ramesh2022hierarchical,saharia2022photorealistic} and image translation \cite{gal2022image}.
Some approaches \cite{kwon2022diffusion,tumanyan2022splicing} consider both structure and appearance in image transfer. Kwon et al.~\cite{kwon2022diffusion} use a diffusion model and a special structural appearance loss for appearance transfer, which performs well in transforming the appearance between similar objects, such as from zebras to horses and from cats to dogs.

However, there are two main challenges when applying the commonly used image transfer methods to the reference-based fashion design task shown in Fig.~\ref{example}. 
First, common image transfer methods only consider the translation between semantically similar images or objects. For example, the transformation in \cite{kwon2022diffusion} is based on the similarity of the semantically related objects in vision transformer (ViT) \cite{amir2021deep} features. In the reference-based fashion design task, the semantic features of reference appearance images are always far different from clothing images. As a result, commonly used image transfer methods usually generate unrealistic fashions in this task and difficult to transfer the appearance. Besides, These methods only transfer the style or appearance, which hardly converts the appearance to a suitable texture material by using a non-clothing image.
Second, image transfer methods \cite{saharia2022palette} usually require a large number of samples from both source and target domains. However, there are no samples available for newly designed output domains, resulting in a lack of guidance during the transfer process. Thus, the generated new fashion images are likely to lose the structural information of the input clothing images.

To address the aforementioned issues, we propose an unsupervised structure-aware transfer framework based on diffusion named \textit{DiffFashion}, which semantically generates new clothes from a given clothing image and a reference appearance image. 
The proposed framework is based on denoising diffusion probabulistic models (DDPM) \cite{ho2020denoising} and preserves the structural information of the input clothing image when transferring the reference appearance with three steps.
First, we decouple the foreground clothing with automatically generated semantic masks by conditioned labels. Then, we encode the appearance image with DDPM which is proven to be the optimal transport process to keep the high-appearance similarity and denoise the image with mask guidance to transfer the structural information. Moreover, we use the ViT for both appearance and structure guidance during the denoising process. This process is illustrated in Fig.~\ref{frame}.

Our contributions are summarized as follows:
\begin{itemize}
    \item We propose a novel structure-aware image transfer framework, which generates structure-preserving fashion designs without knowledge about output domains.
    \item We keep the appearance information by the optimal transport properties of the DDPM encoder.
    \item We employ mask guidance and ViT guidance to transfer structural information in the denoising process.
    \item Extensive experimental results verify that our method achieves state-of-the-art (SOTA) performance in clothing design.
\end{itemize}

The outline of the paper is as follows: In Section~\ref{sec:related_work}, we review state-of-the-art (SOTA) fashion design and image translation methods. Section~\ref{sec:ddpm} introduces the preliminary background of DDPM. We introduce our proposed method in Section~\ref{sec:method}. The experiments of our proposed method are provided in Section~\ref{sec:exp}, followed by the conclusion and future work in Section~\ref{sec:conclude}.
\section{Related Work}
\label{sec:related_work}

\subsection{Fashion Design}
Fashion design models aim to design new clothing from a given clothing collection. Sbai et al. \cite{sbai2018design} use GAN to learn the encoding of clothes, and then use the latent vector to perform the stylistic transformation. Cui et al. \cite{cui2018fashiongan} use the sketch image of the clothes to control the generated structure. 
Good results have been achieved in terms of structural control. 
Yan et al. \cite{yan2022toward} use a patch-based structure to implement texture transfer on generated objects. 
However, they cannot use other images as texture references and their tasks is limited to generating new samples from existing clothes collections.
% Due to the structural problems of GAN, \cite{yan2022toward} can only achieve simple control of color and texture, and cannot use real pictures for reference to automatically complete the design. 
As a result, due to the unreliable training problem of GAN, more advanced methods are needed to achieve improved realism in generated effects.

\subsection{GAN-based Image Transfer}
The image-to-image translation aims to learn the mapping between the source and the target domains, often using a GAN network. Paired data methods like \cite{isola2017image,zhu2017toward} use the target image corresponding to each input for the condition in the discriminator. Unpaired data methods like \cite{dong2017unsupervised,huang2018multimodal,saito2020coco} decouple the common content space and the specific style space in an unsupervised way. But both these methods require amounts of data from both domains. Besides, the encoding structure of GANs makes it difficult to decouple appearance and structural information.  When the gap between the two domains is too large, the result may not be transformed \cite{saito2020coco,zhu2017unpaired,lee2018diverse} or have lost information from the original domain \cite{yang2022unsupervised}.

\subsection{Diffusion Model-based Image Transfer}

Recently, denoising diffusion probabilistic models (DDPMs) have emerged as a promising alternative to GANs in image-to-image translation tasks.
Palette \cite{saharia2022palette} firstly applies the diffusion model in image translation and achieves good results in colorization, inpainting, and other tasks.
% Palette \cite{saharia2022palette} is among the first to apply the diffusion model in image translation, and achieves good results in colorization, inpainting and other tasks. 
However, this approach requires the target image as a condition for diffusion, making it infeasible for unsupervised tasks. 
For appearance transfer, DiffuseIT \cite{kwon2022diffusion} uses the same DINO-ViT guidance as \cite{tumanyan2022splicing}, which greatly improves the realism of the transformation. However, it still cannot solve the problem of lacking matching objects in the clothing design task.
% However, it still cannot address the issue of the lack of matching objects in fashion design tasks.

\subsection{Neural Style Transfer (NST)}

Neural style transfer (NST) has shown great success in transferring artistic styles.
There are mainly two types of approaches to modeling the style or visual texture in NST. One is based on statistical methods \cite{gatys2016image,huang2017arbitrary}, in which the style is characterized as a set of spatial summary statistics. The other is based on non-parametric methods, such as using Markov Random Field \cite{li2016combining,li2016precomputed}, in which they swap the content neural patches with the most similar ones to transfer the style. 
After texture modeling, a pre-trained convolutional neural network (CNN) network is used to complete the style transfer. Although NST-based methods work well for global artistic style transfer, their content/style decoupling process is not suitable for fashion design. 
In addition, NST-based methods assume the transfer is between similar objects or domains.
Tumanyan et al. \cite{tumanyan2022splicing} propose a new NST loss from DINO-ViT, which succeeds in transferring appearance  between two semantically related objects, such as ``cat and dog" or ``orange and ball". However, in our task, there are no specific related objects between the clothing image and the appearance image.

\section{Preliminary of Denoising Diffusion Probabilistic Models }
\label{sec:ddpm}

Diffusion probabilistic models \cite{ho2020denoising,song2020denoising,song2020score} are a type of latent variable model that consists of a forward diffusion process and a reverse diffusion process. In the forward process, we gradually add noise to the data, and then sample the latent  $x_t$ for $t=1,..., T$ as a sequence. Noise added to data in each step is sampled from a Gaussian distribution, and the transmission can be represented as $q(x_t|x_{t-1})=\mathcal{N} (\sqrt{1-\beta_{t}}x_{t-1}, \beta_{t}I)$, where the Gaussian variance $\{\beta_{t}\}_{t=0}^{T}$ can either be learned or scheduled. Importantly, the final latent encoding by the forward process can be directly obtained by,
% \begin{center}
\begin{equation}
\label{eq1}
x_t = \sqrt{\overline{\alpha}_t}x_0 + \sqrt{(1 - \overline{\alpha}_t)}\epsilon,  
     \epsilon \sim \mathcal{N}(0,I), 
\end{equation}
% \end{center}
where $\alpha_t=1 - \beta_t$ and $\overline{\alpha}_t=\prod_{s=1}^t \alpha_s$.
Then in the reverse process, the diffusion model learns to reconstruct the data by denoising gradually. A neural network is applied to learn the parameter $\theta$ to reverse the Gaussian transitions by predicting $x_{t-1}$ from $x_{t}$ as follow:
% \begin{center}
\begin{equation}
\label{eq2}
p_\theta(x_{t-1}|x_t)=\mathcal{N}(x_{t-1};\mu_\theta(x_t,t),\sigma^2I).
\end{equation}
% \end{center}
To achieve a better image quality, the neural network takes the sample $x_t$ and timestamp $t$ as input, and predicts the noise added to $x_{t-1}$ in the forward process instead of directly predicting the mean of $x_{t-1}$. 
The denoising process can be defined as:
% The mean of $x_{t-1}$ is a linear combination of the noise and $x_t$, \textit{i.e.}, 
% The variances $\beta_t$ can be either learned by the neural network or fixed by the noise scheduled in the forward process:
\begin{equation}
    \label{eq3}
    \mu_\theta(x_t,t)=\frac{1}{\sqrt{\alpha_t}}(x_t-\frac{1-\alpha_t}{\sqrt{1-\overline{\alpha}_t}}\epsilon_\theta(x_t,t)),
\end{equation}
where $\epsilon_\theta(x_t,t)$ is the diffusion model trained by optimizing the objective, \textit{i.e.},
\begin{equation}
\label{eq4}
{min}_\theta \mathcal{L}(\theta)=E_{t,x_0,\epsilon}[(\epsilon-\epsilon_\theta(\sqrt{\overline{\alpha}_t}x_0)+\sqrt{1-\overline{\alpha}_t\epsilon},t))^2].
\end{equation}
In the image translation task, there are two mainstream methods to complete the translation. One is using the conditional diffusion model, which takes extra conditions, such as text and labels as input in the denoising process. Then the diffusion model $\epsilon_\theta$ in Eq.~(\ref{eq3}) and Eq.~(\ref{eq4}) can be replaced with $\epsilon_\theta(x_t,t,y)$, where $y$ is the condition.
The other type of method \cite{dhariwal2021diffusion} uses pre-trained classifiers to guide the diffusion model in the denoising process and freezes the weights of the diffusion model. With the diffusion model and a pre-trained classifier $p_\phi(y|x_t)$, the denoising process $\mu_\theta(x_t,t)$ in Eq.~(\ref{eq3}) can be supplemented with the gradient of the classifier, \textit{i.e.}, $\hat{\mu}_\theta(x_t,t)=\mu_\theta(x_t,t) + \sigma_t\nabla log p_\phi(y|x_t)$. 

\begin{figure*}[!t]
\centering
\includegraphics[scale=0.45]{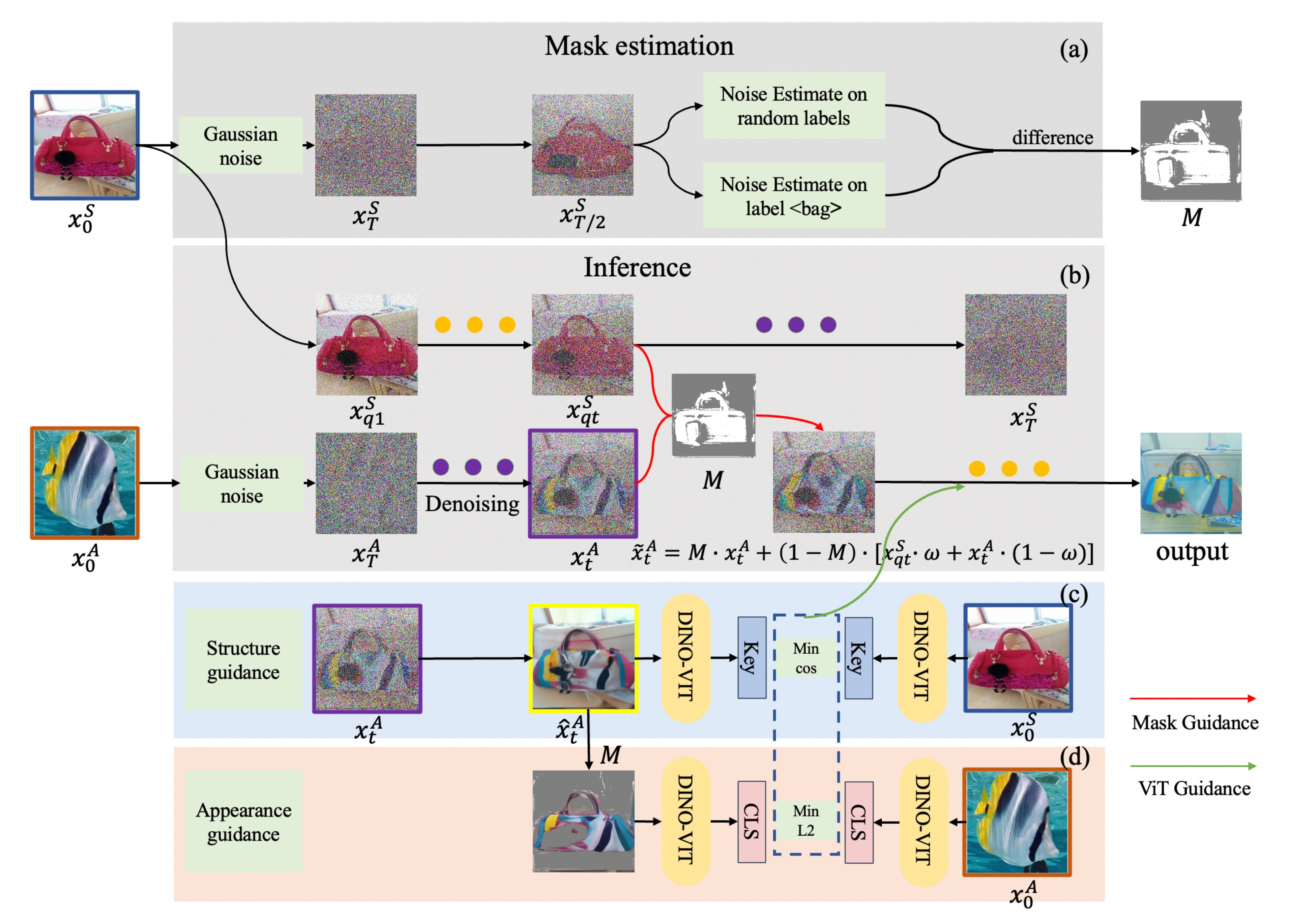}
\caption{The pipeline of our approach. {(a)}:
We add noise to clothing image $x^{S}_0$, and then use different label conditions to estimate the noise in the denoising process. The semantic mask of the $x^{S}_0$ can be obtained from the noise difference. {(b)}: We denoise the reference appearance image $x^{A}_0$. In the denoising process, we use the mask in (a) to replace the background with pixel values obtained from the encoding process at the same timestamp. {(c)} and {(d)}: We use DINO-VIT features to compute structure loss between $x^{A}_t$ and $x^{S}_0$, appearance loss between $x^{A}_t$ and $x^{A}_0$, to guide the denoising process. Purple dots and yellow dots represent the denoising process with the same timesteps respectively.
}
\label{frame}
\end{figure*}
\section{Proposed Method}
\label{sec:method}

\subsection{Overview of Fashion Design with DiffFashion}
% describe task
Given a clothing image $x^{S}_0$ and a reference appearance image $x^{A}_0$, our proposed \textit{DiffFashion} aims to design a new clothing fashion that preserves the structure in $x^{S}_0$ and transfers the appearance from $x^{A}_0$ while keeping it natural, as shown in Fig.~\ref{frame}.
% challenges and issues
We list two main challenges in this task. First, there are no given reference images for the output result since there is no standard answer for fashion design. Without the supervision of the ground truth, it is difficult to train the model. Second, preserving the structure information from the given input clothing image while transferring the appearance is also being under-explored.
% how to solve
To address those two challenges, we present the \textit{DiffFashon}, which is a novel structure-aware transfer model with the diffusion model.
We use the diffusion model \cite{nichol2021improved} pre-trained on Imagenet \cite{deng2009imagenet} for all the denoising processes in DiffFashion. 
% and freeze the pre-trained weight through out the whole procedure.
First, we decouple the foreground clothing with a generated semantic mask by conditioned labels, as shown in Fig.~\ref{frame} (a). Then, we encode the appearance image $x^{A}_0$ with DDPM, and denoise it with mask guidance to preserve the structure information, as shown in Fig.~\ref{frame} (b). Moreover, we use the DINO-ViT \cite{amir2021deep} for both appearance and structure guidance during the denoising process, as shown in Fig.~\ref{frame} (c) and (d). The details are illustrated in the following sections.

% \subsection{Fashion Design with DiffBag}

% Given a source appearance $x_0$ image and a source structure image $y_0$. we transfer the structure of $x_0$ to be similar with $y_0$ , keeping reality and appearance. we use the Diffusion model \cite{nichol2021improved} pre-trained on Imagenet \cite{deng2009imagenet} to complete the image conversion, and freeze the pre-trained weight through out the whole procedure.
% % Fig.\ref{frame} illustrates the framework of our approach, which we detail below.

% First, we generate a semantic mask for the bag by diffusion noise difference estimation, as shown in Fig.\ref{frame} (a). Then we encode the appearance image $x_0$ with DDIM, then denoise it with mask guidance, as shown in Fig.\ref{frame} (b). Besides, we use the ViT for guidance during the denoising process, as shown in Fig.\ref{frame} (c) and (d). 

\subsection{Mask Generation by Label Condition}

To decouple the foreground clothing and background, we generate a semantic mask for the input clothing image $x^{S}_0$ with label conditions. The generated semantic mask is also used for preserving the structure information in later steps.
Existing methods commonly use additional inputs to obtain the foreground region. However, this leads to increased annotation expenses. Inspired by \cite{couairon2022diffedit}, we propose a mask generation approach that can obtain the foreground clothing area without external information or segmentation models. Our approach leverages the label-conditional diffusion model to obtain the desired result.

In the denoising process of the label-conditional diffusion model, there can be different noise estimates for the same latent given negative label conditions like \textit{phone} and \textit{bag}. For these different noise estimates, the regions of the foreground object that are denoised tend to vary little in background regions but greatly in object regions. By taking the difference in the noise area, we can obtain the mask of the object to be edited, as shown in Fig.~\ref{frame}(a).

Instead of generating a mask with the latent of the forward process like \cite{couairon2022diffedit}, we observe that in the denoising process, $x^{S}_t$ has less perceptual appearance information than $x^{S}_{qt}$ (the image in the forward process with timestamp $t$). Therefore, we generate a mask from the image in the denoising process $x^{S}_{t}$ instead of the image $x^{S}_{qt}$ in the forward process. Although the structure of $x^{S}_t$ may have some slight variations, it still provides a better representation of the overall structure information of the foreground object.

Specifically, we input the clothing image $x^{S}_0$ into the diffusion model. After DDPM encoding in the forward process, we obtain the image latent $X^{S}_{T/2}$ in half of the reverse process. 
Denote the foreground label as $y_p$, representing the foreground clothing object. 
% Denote non-foreground labels as $y_n$, representing negative objects. 
% Using the true label, such as ``bag", we first generate a noise map:
Then the noise map for the foreground clothing can be obtained by
\begin{equation}
\label{eq7}
    M_{p} = \epsilon_\theta(\hat{x}^{S}_{T/2},T/2,y_p),
\end{equation}
where $\hat{x}^{S}_{T/2}$ is the estimated source image predicted from $x^S_{T/2}$ by Tweedie's method \cite{kim2021noise2score}, \textit{i.e.},
\begin{equation}
    \hat{x}_t = \frac{x_{T/2}}{\sqrt{\Bar{\alpha}_{T/2}}} - \frac{\sqrt{1-\Bar{\alpha}_{T/2}}}{\sqrt{\Bar{\alpha}_{T/2}}}{ \epsilon}_\theta(x_{T/2},T/2,y_p).
\label{eqadd1}
\end{equation}
Denote non-foreground labels as $y_n$, representing negative objects.
We use $N$ different non-foreground label conditions to get an averaged noise map, \textit{i.e.},
\begin{equation}
\label{eq8}
    M_{n} = \frac{1}{N} \sum_{i=1}^N\epsilon_\theta(\hat{x}^{S}_{T/2},T/2,y_{i}),
\end{equation}
where $i \in \{1, ..., N\}$.
The difference between the two noise maps $M_p$ and $M_n$ can be obtained. Then we set a threshold for binarization, which returns an editable semantic mask $M$ for the foreground clothing region.

\subsection{Mask-guided Structure Transfer Diffusion}

It is difficult to transfer the appearance of the original image to a new fashion clothing image when the gap between the two domains is too large \cite{tumanyan2022splicing}. Because such methods control the appearance by a single loss of guidance, the redundant appearance information of the structure clothing reference image cannot be completely eliminated. Besides, when using a natural non-clothing image for appearance reference, the generated texture maybe not be suitable for clothing. Because these models only transfer the style or appearance. The appearance cannot be converted to a suitable texture material like cotton for clothing.
In DiffFashion, to address this problem, rather than transferring from the input clothing image $x^{S}_0$, we transfer from the reference appearance image $x_0^{A}$ to the output fashion clothing image with the guidance of the structural information of the input clothing image.

% Inspired by \cite{khrulkov2022understanding}, which proves that the forward procedural encoder of DDPM performs an optimal transfer from the natural distribution to $\mathcal{N}(0,I)$ using $l_2$ loss. This is demonstrated by the spatial high-level texture similarity of the image. Experiments have shown that for the same DDPM encoding latent with different label conditions used for denoising, the resulting  natural images have similar textures and semantic structures.

Inspired by \cite{khrulkov2022understanding}, it has been shown that for the same DDPM encoding latent with different label conditions used for denoising, the resulting  natural images have similar textures and semantic structures.
% Instead using the DDPM latent $X_T$ obtain from the structure image to keep the structure information. 
We use the latent $x^{A}_t$ of the reference appearance image to transfer more appearance information to the output fashion. Besides, the texture of the appearance image can be transferred more realistic and suitable for clothing in the denoising process. Meanwhile, the semantic mask $M$ obtained from the previous step is used to preserve the structure of the clothing image.
As shown in Fig.~\ref{frame}{(b)}, the appearance image $x^{A}_0$ is first used to encode by the forward process of DDPM. Then the mask-guided denoising process is employed. 

Specifically, at each step in the denoising process, we estimate the new prediction $x^{A}_t$ from the diffusion model as follows, 
\begin{equation}
\label{eq9}
x^{A}_t=\frac{1}{\sqrt{\alpha_{t+1}}}(x^{A}_{t+1}-\frac{1-\alpha_{t+1}}{\sqrt{1-\overline{\alpha}_{t+1}}}\epsilon_\theta(x^{A}_{t+1},t+1,y_p)).
\end{equation}
Then we combine the transferred foreground appearance $x^{A}_t$ and the clothing image of corresponding timestamp $x^{S}_{qt}$ with the generated mask $M$ as guidance, \textit{i.e.}, 
% We keep the pixels in the mask, and for the pixels outside the mask, we use the bag image with the pixels of corresponding timestep $x_{qt}$ to mix with $x^{\prime}_t$. 
\begin{equation}
\label{eq10}
    \tilde{x}^{A}_t = M \cdot  x^{A}_t + (1 - M)\cdot[\omega_{mix} \cdot x^{S}_{qt} + (1 - \omega_{mix}) \cdot x^{A}_t],
\end{equation}
where $\omega_{mix}$ is the mix ratio of the appearance image and the clothing image.
This change ensures that the appearance information in the mask is transferred, while other structural information keeps consistent with the clothing image.

%Due to the uncontrollable effect of the mask, and when the structural gap between the two objects is too large, the information in $x$ may be removed by the mask. We perform mask guidance in partial steps. 
%To mitigate the uncontrollable effect of the mask and avoid information loss when the structural gap between the two objects is too large, we implement the mask guidance in partial denoising steps to prevent the information in $x$ being removed by the mask.

\subsection{ViT Feature Guidance}
As mentioned in \cite{tumanyan2022splicing,kwon2022diffusion}, the structure features and appearance features can be separated with DINO-ViT \cite{amir2021deep}. We use both appearance guidance and structure guidance in the denoising process to keep the output image realistic. 

Following \cite{tumanyan2022splicing,kwon2022diffusion}, we employ the $[CLS]$ tokens in the last layer of ViT to guide the semantic appearance information as follows,
\begin{equation}
\begin{aligned}
\label{eq12}
& \mathcal{L}_{app}(x^{A}_0,\hat{x}^{A}_t) = \\
& ||e_{[CLS]}^L(x^{A}_0)-e_{[CLS]}^L({\hat{x}^{A}_t})||_2 + \lambda_{MSE}||x^{A}_0-{\hat{x}^{A}_t}||_2,
\end{aligned}
\end{equation}
where $e_{[CLS]}^L$ is the last layer $[CLS]$ token, and $\lambda_{MSE}$ is the coefficient of global statistic loss between images.
%With the object semantic mask $M$. 
To better leverage the appearance between the object and the appearance image,
we use the object semantic mask $M$ to remove the background pixel of $\hat{x}^{A}_t$ in Eq.~\ref{eq12}, and only compute the appearance loss of the object within the mask.

In addition, we adopt a patch-wise method in the structural loss to better leverage the local features. We adopt the $i$-th key vector in the $l$-th attention layer of the ViT model, denoted as $k_i^l(x_t)$, to guide the structural information of the $i$-th patch of the original clothing image as follows,
% In order to achieve more precise structural similarities, especially the design of the stripe areas inside the clothes, we refer to the designs of \cite{tumanyan2022splicing} and \cite{kwon2022diffusion}. They separate the structure features and appearance features with a DINO-ViT. Briefly, their experiments show that in ViT, the $[CLS]$ tokens in the last layer contain semantic appearance information. At the same time, the $k_i(x_t)$ vector in the attention layer can express the structural information of the original image well. Inspired by neural style transfer, where a loss is designed that can measure the structural similarity and appearance similarity of two pictures, \cite{kwon2022diffusion} further improves the local feature of this loss with a patch-wise method as follows:
\begin{equation}
\begin{aligned}
    \label{eq11}
    & \mathcal{L}_{struct}(x^{A}_0,\hat{x}^{A}_t) = \\
    & -\sum_{i}\log\left(\frac{\mathrm{sim}(k_i^{l,S}, k_i^{l,A})}{ \mathrm{sim}(k_i^{l,S}, k_i^{l,A}) + \sum_{j\neq i} \mathrm{sim}(k_i^{l,S}, k_j^{l,A})}\right),
\end{aligned}
\end{equation}
where $\mathrm{sim}(\cdot,\cdot)$ is the exponential value of normalized cosine similarity, \textit{i.e.},
\begin{equation}
\label{eq13}
    \mathrm{sim}(k_i^{l,S}, k_j^{l,A}) = \exp \left(\mathrm{cos}\left({k}^l_i({x}^{S}_0), {k}^l_j(\hat{x}^{A}_{t})\right)/\tau \right),
\end{equation}
and $\tau$ is the temperature parameter. 
% Since the key $k^l_i$ contains the spatial information corresponding the i-th patch location in ViT's l-th self-attention layer. 
By using the loss in Eq.~(\ref{eq11}), we minimize the loss between keys at the same position of two images while maximizing the loss between keys of different positions.
% And for appearance loss guidance, we use both ViT guidance and normal L2 loss guidance as follow:
Then our total loss for guidance as follow:
\begin{equation}
\begin{aligned}
    \label{eq14}
    \mathcal{L}_{total} = \lambda_{struct}\mathcal{L}_{struct}+ \lambda_{app}\mathcal{L}_{app}, 
\end{aligned}
\end{equation}
where $\lambda_{struct},\lambda_{app}$ are the coefficient of structure loss and appearance loss.

\section{Experiments}
\label{sec:exp}

In this section, we describe our fashion design dataset and experiment settings. We also demonstrate the qualitative and quantitative results to show the effectiveness of our proposed method.
\subsection{Dataset} 
\begin{figure*}[!t]
\centering
\includegraphics[scale=0.65]{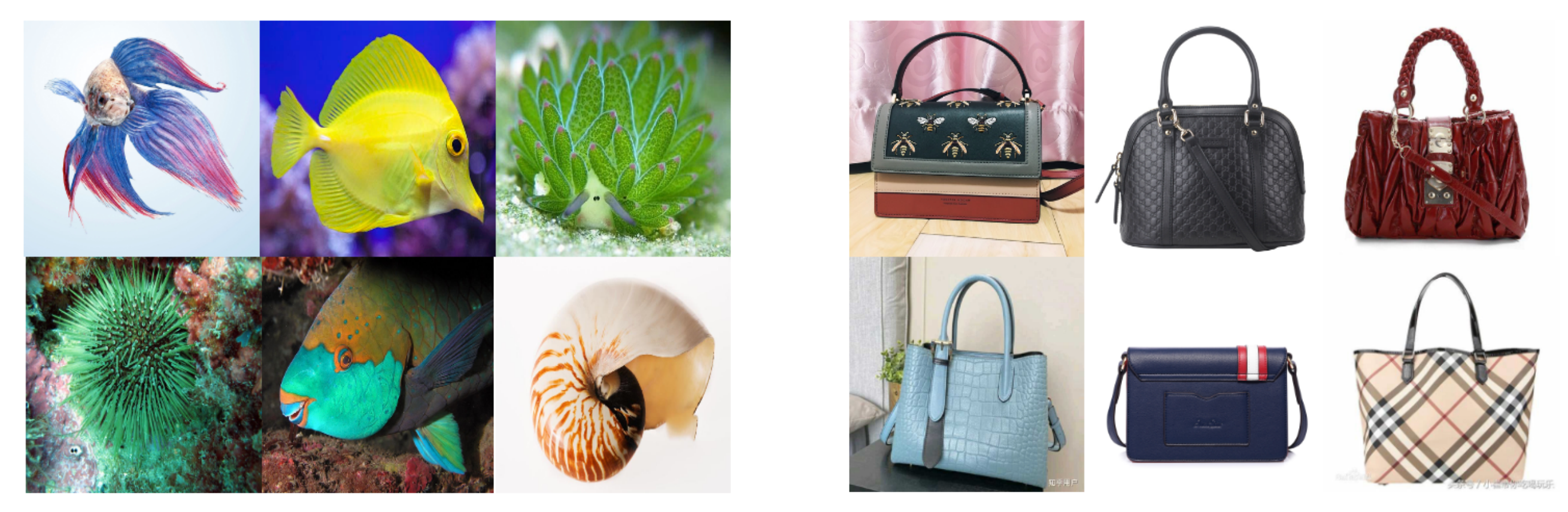}
\caption{Samples from our proposed dataset of \textit{OceanBag}. The left part shows some examples of marine life images, and the right part shows some samples of bag images.}
\label{datasetfig}
\end{figure*}

To our best knowledge, there is no specific reference-based fashion design dataset currently. Thus, we collect a new dataset, namely \textit{OceanBag}, with real handbag images and ocean animal images as reference appearances for generating new fashion designs. 
% So we conduct experiments on dataset collected by ourselves, which mainly consists of real bag pictures and real ocean animals pictures. 
\textit{OceanBag} has 6,000 photos of handbags in various scenes and 2,400 pictures of various marine lives in the real world, among various marine scenes such as fish swimming on the ocean floor. The 2,400 marine scene images contain more than 80 kinds of marine organisms, 50\% of which are fish, as well as starfish, crabs, algae, and other sea creatures, as shown in Fig.~\ref{datasetfig}. 
In our experiments, we screened 30 images for experiments based on diversity such as background complexity, species, and quantity of organisms.

\begin{table}[!t]
\begin{center}
\caption{Overall information about the \textit{OceanBag} dataset.}
\begin{tabular}{l| c c c }
\toprule

         Dataset & Quantity & Image size & Complex ratio \\
         \midrule
          Handbag  &  6,000     &  256$\times$256 & 0.16      \\
          Marine life  &  2,400    &   256$\times$256 & 0.43 \\
\bottomrule
\end{tabular}

\label{datasettable}
\end{center}
\end{table}

\begin{figure*}[!t]
\centering
\includegraphics[scale=0.65]{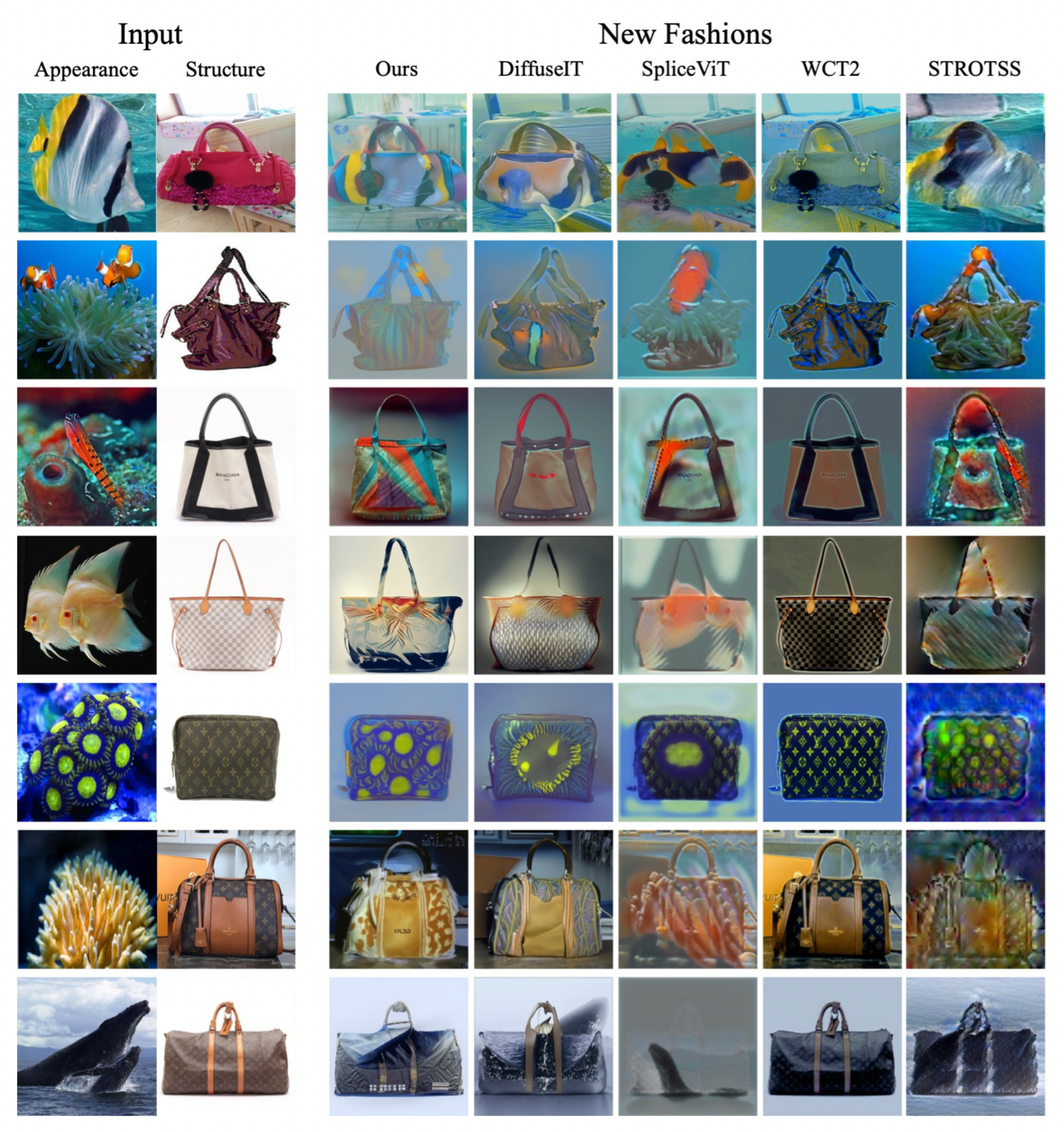}
\caption{Comparison with other state-of-the-art (SOTA) methods. Our results show better performance in both appearance and structure similarity.}
\label{res}
\end{figure*}

\begin{table}[!t]
\begin{center}
\caption{Results of the user study. The output fashion images are evaluated based on their realism, structure, and appearance scores, ranging from 0 to 100. The overall performance is the average of the three scores. The best performance is shown in bold and the second best is shown in light blue.}
\begin{tabular}{L{1.4cm}|C{0.7cm}C{1.5cm}C{1.5cm}C{1.5cm}}%{l| c c c c }
     	\toprule
         Method & \textbf{Overall} & Realism & Structure & Appearance\\
         \midrule
          DiffuseIT  & \textcolor{cyan}{67.09} & $75.53 \pm 4.68$     &  $88.46  \pm 5.40$ & $37.27  \pm 8.36$  \\
          SpliceViT & 60.98 & $68.44  \pm 4.36$ &   $80.80  \pm 7.82$ & $33.70  \pm 8.32$  \\ 
          WCT2 & 65.45 & $ \textbf{82.89}  \pm 5.42$ &$ \textbf{95.76}  \pm 1.84$ & $17.69  \pm 5.77$     \\
          
          STROTSS & 63.00 & $63.33  \pm 6.43$ &   $82.55  \pm 6.65$ & $\textcolor{cyan}{43.11}  \pm 8.93$  \\
           \textbf{Ours} & \textbf{75.04} &  $\textcolor{cyan}{81.15}  \pm 4.76$     &  $\textcolor{cyan}{91.07}  \pm 3.82$ &  $\textbf{52.89}  \pm 7.92$  \\
          \bottomrule
\end{tabular}

\label{res_table}
\end{center}
\end{table}

\begin{table}[!t]
\begin{center}
\caption{Evaluation results based on other models. The best performance is shown in bold and the second best is shown in light blue. ``C.loss", ``M.recall" and ``M.prec." represent classification loss, Mask-RCNN recall, and Mask-RCNN precision, respectively.}
\begin{tabular}{l| c c c c}
     	\toprule
         Method & C.loss & M.recall & M.precision & CDH \\
          \midrule
          DiffuseIT &$7.62  \pm 4.38$
           & 0.17 &0.16& 0.13 \\

          SpliceViT &$\textcolor{cyan}{6.03} \pm 3.68$ &  0.03&0.03&0.07\\
          WCT2 &$9.56  \pm 4.16$  & \textbf{0.53}  &\textbf{0.51}&\textcolor{cyan}{0.23} \\
          STROTSS &$11.20 \pm 3.52$& 0.06 &0.06& \textbf{0.43}\\
          \textbf{Ours}   &  $\textbf{5.93} \pm 4.72$     &  \textcolor{cyan}{0.2}& \textcolor{cyan}{0.17}& 0.22
          \\
          \bottomrule
          
\end{tabular}
\label{res_cls}
\end{center}
\end{table}

\begin{figure*}[!t]
    \centering
    \includegraphics[width=1\linewidth]{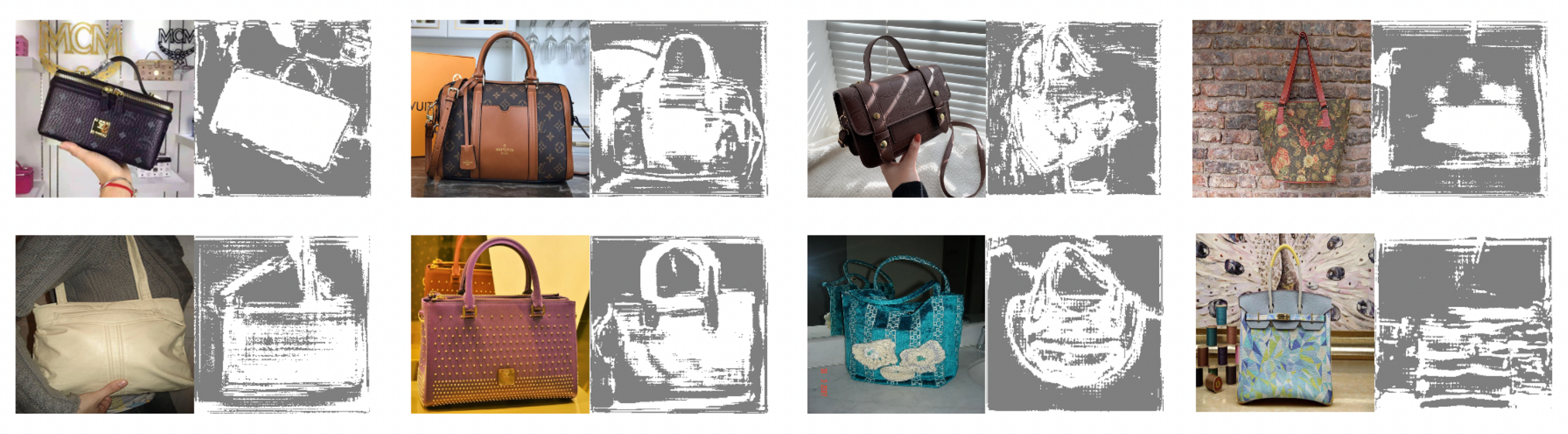}
    \caption{Illustration of Mask Generation by Label Condition.}
    \label{fig:Mask Generation}
\end{figure*}

\begin{figure*}[!t]
    \centering
    \includegraphics[width=0.8\linewidth]{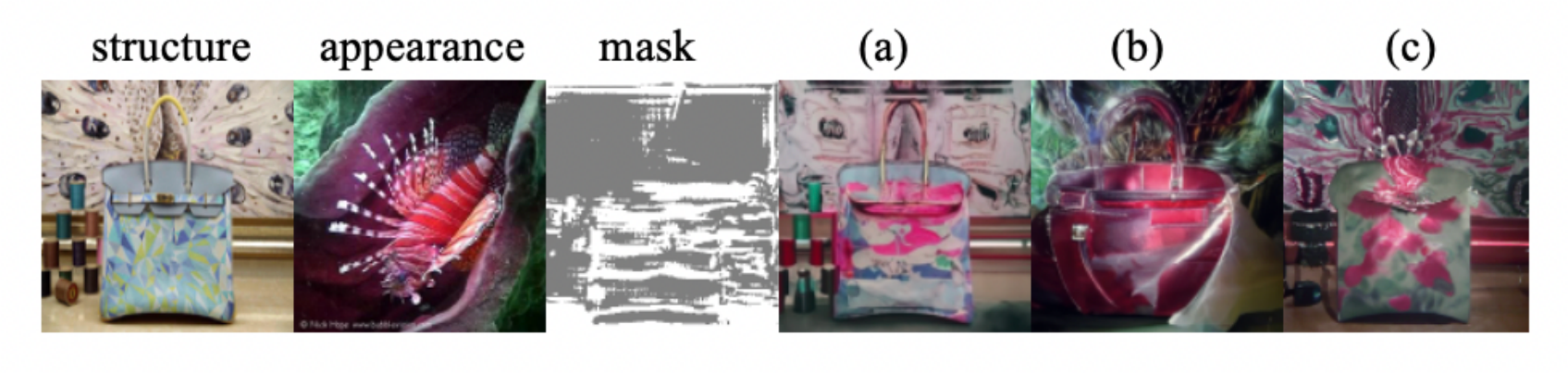}
    \caption{An example of fashion output with a generated messy mask. (a) and (b) are our results with and without mask guidance, respectively. (c) is the result of DiffuseIT.}
    \label{fig:my_label}
\end{figure*}

\begin{figure*}[!t]
    \centering
    \includegraphics[width=0.65\linewidth]{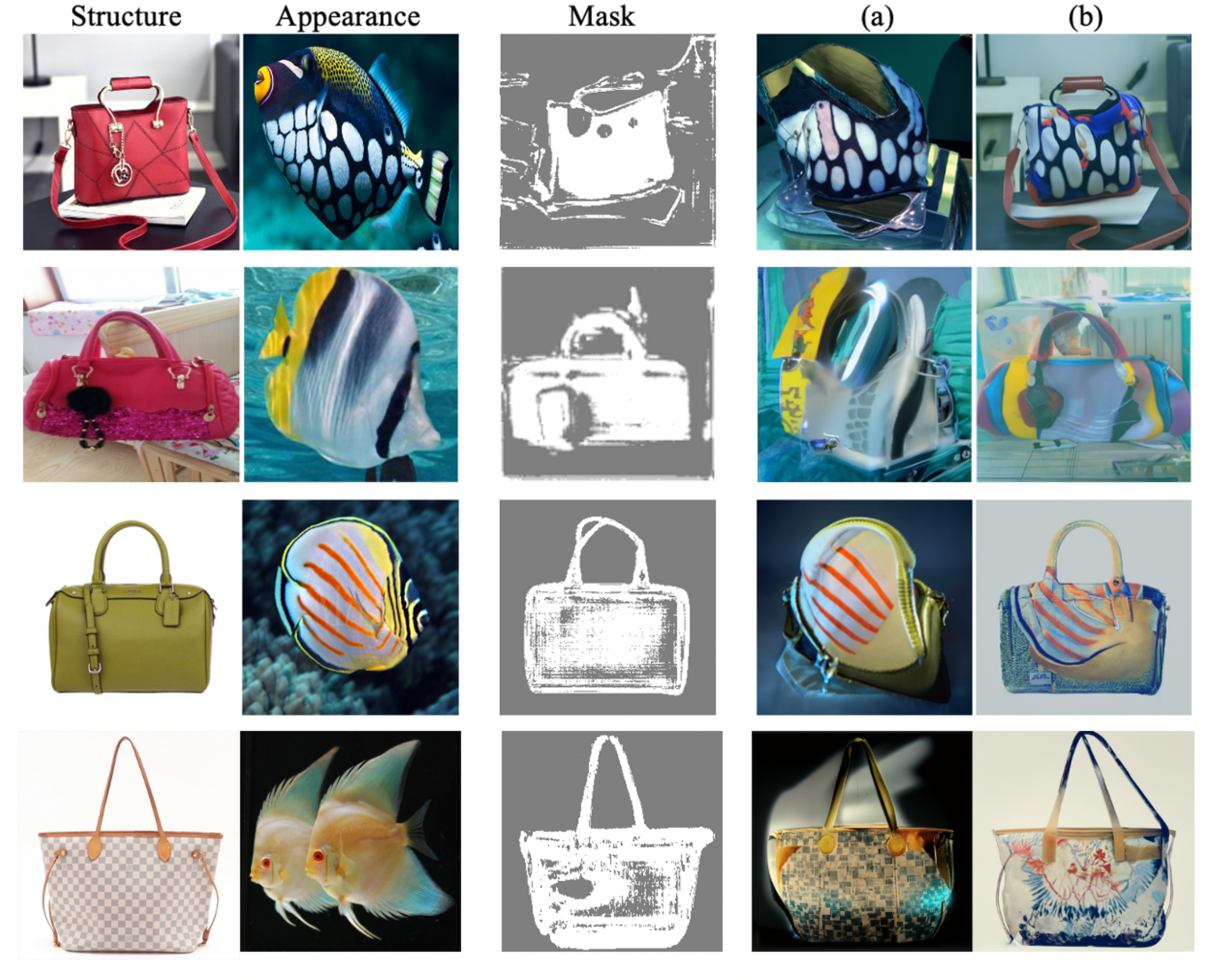}
    \caption{Examples that show the mask effectiveness. {(a)} and {(b)} show the results of our method with or without mask guidance, respectively}
    \label{fig:Mask Effectiveness}
\end{figure*}

\begin{figure}[!t]
    \centering
    \includegraphics[width=1\linewidth]{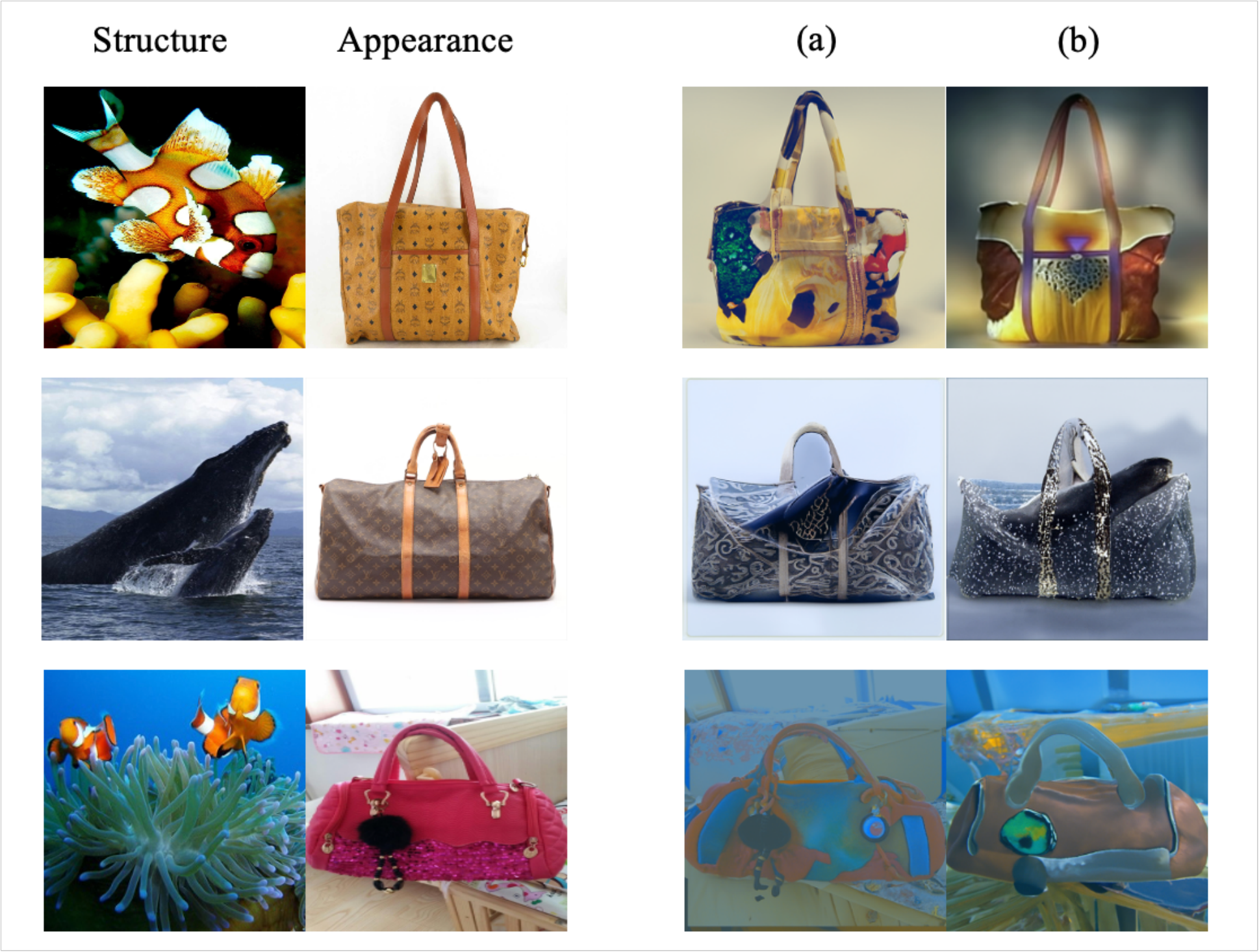}
    \caption{Comparison with label-conditional DiffuseIT. Our results and results from DiffuseIT with label-conditional diffusion models are shown in (a) and (b), respectively.}
    \label{fig:Comparition}
\end{figure}

\begin{figure}[!t]
    \centering
    \includegraphics[width=1\linewidth]{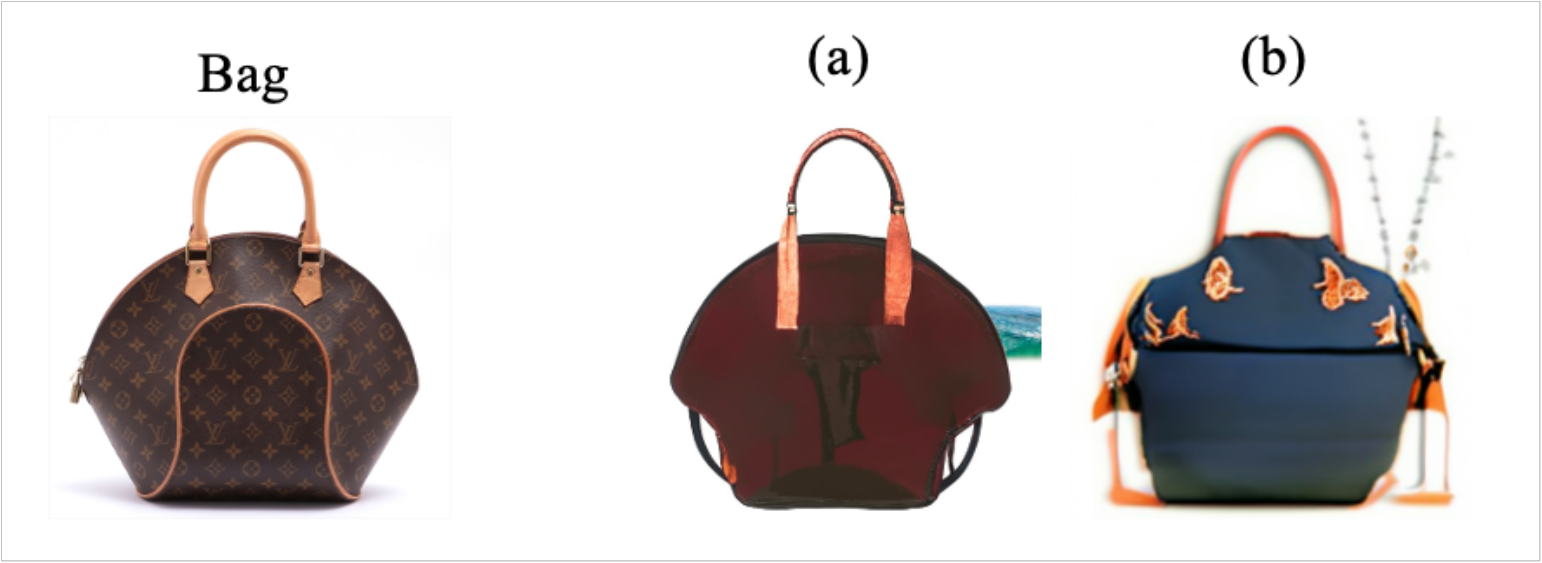}
    \caption{Examples of the DiffuseIT model with the text guidance. ``Handbag" to ``Handbag with marine life pattern" and ``Ocean style Handbag" are prompts for (a) and (b), respectively. }
    \label{text}
\end{figure}

We refer to images with solid backgrounds as simple backgrounds, while those with real scenes are referred to as complex backgrounds. The complex ratio in Table~\ref{datasettable} shows the proportion of complex background images in the dataset. The complex background of the marine biological dataset is usually real ocean pictures such as the seabed and the deep sea. For the bag images in the dataset, the complex backgrounds often include scenes of mall containers or tables. 
% For convenience, we annotate the bounding box for the bag dataset.

\subsection{Experimental Setup}
\label{setup}
We conduct all experiments using a label-conditional diffusion model \cite{nichol2021improved} pre-trained on the ImageNet dataset \cite{deng2009imagenet} with $256\times256$ resolution. In all experiments, we use a diffusion step of $T$ = 60 and re-sampling repetitions of $N$ = 10. In a single RTX 3090 unit, it takes 20 seconds to generate each mask and 120 seconds to generate each image. For fairness of comparison, other parameters in the diffusion model are kept the same as \cite{kwon2022diffusion}.

In the mask generation part, we set the binarization threshold to -0.2. Due to the stochastic nature of the diffusion model, we generate masks using three different sets of labels, including ``cellphone, forklift, pillow", ``waffle iron, washer, guinea pig" and ``brambling, echidna, custard apple". Then we choose the best one among them for guidance. To ensure a fair comparison, We run the baseline DiffuseIT \cite{kwon2022diffusion} three times as ours.

In the guidance part, to mitigate the uncontrollable effect of the mask and avoid information loss when the structural gap between the two objects is too large, we use mask guidance in the first 50\% steps of the denoising stage, and the mix ratio $\omega_{mix}$ is set to 0.98. In the ViT guidance part, we set the coefficient of appearance loss $\lambda_{app}$ to 0.1 and 1 for structure loss $\lambda_{struct}$. And we keep other parameters the same as DiffuseIT \cite{kwon2022diffusion}.

\subsection{Evaluation Methods and Metrics} 
There is currently no existing automatic metric suitable for evaluating fashion design across two natural images. To keep the fashion image realistic, the migration degree of the appearance and the similarity of the structure sometimes are mutually contradictory when measured. To compare among different methods, we follow existing appearance transfer/fashion design works \cite{tumanyan2022splicing,kwon2022diffusion,jing2019neural,kim2020deformable,mechrez2018contextual,park2020swapping}, which rely on human perceptual evaluation to validate the results. 
% And we also show our results for qualitative comparison.

\subsection{Experimental Results}
We perform both quantitative and qualitative evaluations on the \textit{OceanBag} dataset. We compare our model with SplicingViT \cite{tumanyan2022splicing}, DiffuseIT \cite{kwon2022diffusion} , WCT2 \cite{yoo2019photorealistic} and STROTSS \cite{kolkin2019style}. 
Fig.~\ref{res} shows qualitative results for all methods. In all examples, it can be seen that in terms of fashion design, our method has achieved better performances in terms of realism and structure, while completing appearance transfer. 
As for the DINO-ViT-based image-to-image translation methods, DiffuseIT successfully keeps the structure for most images, but it shows less appearance similarity. SplicingViT transfers the appearance well, but its results are far away from realistic fashion images.
NST methods like STROTSS and WCT2 effectively retain the structure of the source image, but WCT2 outputs exhibit limited changes apart from color adjustments. Although STROTSS successfully transfers the appearance, its results often suffer from color bleeding artifacts and thus show less authenticity.

We also conduct a user study to evaluate the samples and obtain subjective evaluations from participants. Specifically, we ask 30 users to score all the output fashion images from all methods for each input pair. Detailed questions we have asked are as follows: 1) Is the picture realistic? 2) Is the image's structure similar to the input image? 3) Is the output appearance similar to the input appearance image? The scores range from 0 to 100. The overall score is the average of the three scores.
We show the averaged subjective evaluation results in Table~\ref{res_table}. Our model obtains the best score in the overall performance and appearance correlation, and the second place in structure similarity and realism. WCT2 shows the best in realism and structure similarity scores, but it shows the worst score in appearance correlation because the outputs are almost unchanged from the inputs except for the overall color.
Both the qualitative and subjective evaluations show the effectiveness of our proposed method.

Following \cite{tumanyan2022splicing}, we also adopt other pre-trained models to evaluate the result.
% Due to the lack of indicators for our task, we refer to the setting of \cite{tumanyan2022splicing}, and use other pre-trained models to evaluate the result. 
We use the classifier pre-trained with the ImageNet dataset given by improved DDPM \cite{nichol2021improved} and calculate the average classification loss. We also apply Mask-RCNN pre-trained on the COCO dataset to detect the mask of the object of each method. The results are shown in Table~\ref{res_cls}. Our model achieves the lowest classification loss. At the same time, since Mask-RCNN is trained on out-of-distribution (OOD) data, the overall recall rate is quite low. Our model demonstrates the second-best performance after WCT2, but WCT2 only transforms the color for the whole image. Besides, we calculate the color difference histogram (CDH) \cite{liu2013content} between the result and appearance image for each method. Our method achieves better appearance similarity than image translation methods. Although NST methods like STROTSS have a better CDH, they tend to transfer the whole image with simple color transformation, as shown in Fig.~\ref{res}.

\subsection{Ablation Study}

% \begin{figure}[!t]
% \begin{minipage}[t]{1.0\linewidth}
% \vspace{0pt}
%     \centering
%     {\includegraphics[width=1\linewidth]{figs/ab7.png}\caption{\label{fig:Mask Effectiveness}Mask Effectiveness.}}
%     % \vspace{2pt}
%     {\includegraphics[width=\linewidth]{figs/ab5.png}\caption{\label{fig:Mask Generation}Mask Generation.}}
% \end{minipage} %\par
% \medskip
% \begin{minipage}[t]{.5\linewidth}
% \vspace{0pt}
%     \centering
%     {\includegraphics[width=1\linewidth]{figs/ab3.png}\caption{\label{text}DiffuseIT text guidance part.}}
    
%     {\includegraphics[width=\linewidth]{figs/ab4.png}\caption{\label{fig:Comparition}Comparition with label-conditional DiffuseIT.}}
% \end{minipage}
% \end{figure}

In order to verify the effectiveness of the method, we study the individual components of our technical designs through several ablation studies as illustrated from Fig.~\ref{fig:Mask Generation} to Fig.~\ref{text}.

\subsubsection{Mask Generation} We randomly select several bag images with backgrounds from ImageNet and our dataset. % We use three different sets of labels to obtain three different semantic masks and show the best one in
We keep the same experimental setup as Section~\ref{setup} and show the masks in Fig.~\ref{fig:Mask Generation}. For most images, it can generate a foreground object mask that is suitable for our models. Due to the randomness of diffusion, in the last column, we show the scene where the mask is not good enough. But even so, our model still outperforms other models, as shown in Fig.~\ref{fig:my_label}.

\subsubsection{Mask Guidance}
We conduct an experiment on our model without the mask guidance part, as shown in Fig.~\ref{fig:Mask Effectiveness}. Fig.~\ref{fig:Mask Effectiveness}(a) shows the result without mask guidance and Fig.~\ref{fig:Mask Effectiveness} (b) presents the outputs of our model with mask guidance. Without mask guidance, in many images, the structure of the bag is destroyed during diffusion. In the last row of the figure, we show that for some images, using a mask may reduce the correlation of appearance, but this is still enough to complete the transfer task. In order to solve a small number of such problems, we set the probability of 0.2 when applying without using mask guidance.

\subsubsection{Label-Condition} Because our model uses the diffusion model with label-condition, for a fair comparison, we replace the diffusion model of diffuseIT with the same model as ours and use the label ``bag'' for the condition in the denoising stage.
Fig.~\ref{fig:Comparition}(a) shows the results of DiffuseIT with label condition, and Fig.~\ref{fig:Comparition}(b) presents our method. Our method still shows better results in structure preservation, appearance similarity, and authenticity.
In addition, We show some results of a multi-modal guided diffusion model trained on the same amount of data. 
Fig.~\ref{text} shows the result of DiffuseIT with the text guidance. 
``Handbag" to ``Handbag with marine life pattern" and ``Ocean style Handbag" are prompts for (a) and (b), respectively. 
We can see that a text-guided model cannot complete the task well.

\section{Conclusion and future work}
\label{sec:conclude}

% \noindent\textbf{Conclusion} 
We tackle a new problem set in the context of fashion design: designing new clothing fashion from a given clothing image and a natural appearance image, and keeping the structure of the clothing with a similar appearance to the natural image. We propose a novel diffusion-based image-to-image translation framework by swapping the input latent with structure transfer. And the model is guided by an automatically generated foreground mask and both structure and appearance information from the pre-trained DINO-ViT model.
% And we further proposed a structure transfer guidance method by using the mask automatically obtained from the diffusion model. 
The experimental results have shown that our proposed method outperforms most baselines, demonstrating that our method can better balance authenticity and structure preservation while also achieving appearance migration.
% \noindent\textbf{Limitation and Future Work} 
% Although we reduce the dependence of the performance on the mask by using soft mask guidance. However, due to the randomness of diffusion, the mask cannot guarantee good results every time. And for some pictures, due to the high similarity with the background, we cannot obtain a better mask, which will lead to insignificant appearance migration.
Due to the randomness of diffusion, the mask cannot guarantee good results every time. In the future, we will try to constrain the diffusion model using the information condition of other modalities to generate better masks. 

\section*{Acknowledgments}
This work is supported by National Natural Science Foundation of China (62106219) and Natural Science Foundation of Zhejiang Province (QY19E050003).

% \newpage

% \section*{Acknowledgments}
% This should be a simple paragraph before the References to thank those individuals and institutions who have supported your work on this article.

% {\appendix[Proof of the Zonklar Equations]
% Use $\backslash${\tt{appendix}} if you have a single appendix:
% Do not use $\backslash${\tt{section}} anymore after $\backslash${\tt{appendix}}, only $\backslash${\tt{section*}}.
% If you have multiple appendixes use $\backslash${\tt{appendices}} then use $\backslash${\tt{section}} to start each appendix.
% You must declare a $\backslash${\tt{section}} before using any $\backslash${\tt{subsection}} or using $\backslash${\tt{label}} ($\backslash${\tt{appendices}} by itself
%  starts a section numbered zero.)}


\bibliographystyle{elsarticle-num}
\bibliography{IEEEabrv,ref.bib}

\vfill

\end{document}